\pdfoutput=1
\documentclass[11pt,a4paper]{article}
\usepackage[hyperref]{emnlp2020}
\usepackage{times}
\usepackage{latexsym}
\usepackage{soul}
\usepackage{url}
\usepackage{graphicx}
\usepackage{amsmath}
\usepackage{booktabs}
\usepackage{latexsym}
\usepackage{microtype}
\usepackage{subfigure}
\usepackage{graphicx}
\usepackage{amssymb}
\usepackage[boxed,ruled,commentsnumbered]{algorithm2e}
\usepackage{multirow}
\usepackage{makecell}
\usepackage{verbatim}
\usepackage{changepage}

\usepackage{microtype}

\aclfinalcopy 

\title{FedNER: Privacy-preserving Medical Named Entity Recognition with Federated Learning}

\author{Suyu Ge$^1$, Fangzhao Wu$^2$, Chuhan Wu$^1$, Tao Qi$^1$, Yongfeng Huang$^1$ and Xing Xie$^2$\\
  $^1$Department of Electronic Engineering, Tsinghua University, Beijing 100084, China  \\
  $^2$Microsoft Research Asia\\
  {\tt \{gesy17,wu-ch19,qit16,yfhuang\}@mails.tsinghua.edu.cn}\\
  {\tt \{fangzwu,xing.xie\}@microsoft.com} \\}

\date{}

\begin{document}
\maketitle

\begin{abstract}

Medical named entity recognition (NER) has wide applications in intelligent healthcare.
Sufficient labeled data is critical for training accurate medical NER model.
However, the labeled data in a single medical platform is usually limited.
Although labeled datasets may exist in many different medical platforms, they cannot be directly shared since medical data is highly privacy-sensitive.
In this paper, we propose a privacy-preserving medical NER method based on federated learning, which can leverage the labeled data in different platforms to boost the training of medical NER model and remove the need of exchanging raw data among different platforms.
Since the labeled data in different platforms usually has some differences in entity type and annotation criteria, instead of constraining different platforms to share the same model, we decompose the medical NER model in each platform into a shared module and a private module.
The private module is used to capture the characteristics of the local data in each platform, and is updated using local labeled data.
The shared module is learned across different medical platform to capture the shared NER knowledge.
Its local gradients from different platforms are aggregated to update the global shared module, which is further delivered to each platform to update their local shared modules.
Experiments on three publicly available datasets validate the effectiveness of our method.

\end{abstract}

\section{Introduction}

Medical named entity recognition (NER) aims to identify medical entities (e.g., drug names, adverse reactions and symptoms) from unstructured medical texts and classify them into different categories~\cite{tang2013recognizing}.
It can be used in many intelligent healthcare tasks such as pharmocovigilance and health monitoring~\cite{wang2013rational}.
Medical NER has attracted increasing attentions in NLP community, and many methods have been proposed~\cite{alex2007recognising,ekbal2013stacked,dai2017medication}.
For example, Habibi et al.~\shortcite{habibi2017deep} proposed a LSTM-CRF approach, which used LSTM to encode context information within a sentence and used CRF to jointly decode word labels.
Gridach~\shortcite{gridach2017character} further improved this approach by adding an additional character-level LSTM to better encode medical words.

Sufficient labeled data is critical for these methods to train accurate medical NER model~\cite{ratinov2009design}.
However, the labeled medical data in a single medical platform such as a hospital is usually limited.
Annotating sufficient labeled data for medical NER is very expensive and time-consuming, and requires huge expertise in medical domain~\cite{abacha2011medical}.
Although many medical platforms may have some annotated medical NER datasets, they cannot be directly shared to train medical NER models since medical data has rich information of patients and is highly privacy-sensitive.

Recently, McMahan et al.~\shortcite{mcmahan2017communication} proposed a privacy-preserving machine learning framework named federated learning, where the user data is locally stored and a master server coordinates massive user devices to collaboratively train a global model by aggregating the local model updates.
Motivated by federated learning, in this paper we propose a privacy-preserving medical NER method named \textit{FedNER}.
It can leverage the knowledge in the labeled data of different medical platforms to boost the training of medical NER model in each platform without uploading or exchanging the raw medical data.
Since the labeled data in different platforms may have some differences in entity type and annotation criteria,
different from the original federated learning framework where all users share the same model, in our \textit{FedNER} method we decompose the medical NER model in each platform into a shared module and a private module.
The private module is used to capture the characteristics of the local data in each platform, and is updated using the gradients computed from the local labeled data.
The shared module is used to capture the shared knowledge among different platforms to empower the training of medical NER model in each single platform.
Its gradients from different medical platforms are aggregated into a unified one to update the global shared module, which is further delivered to each platform to update the local shared module.
Above process is repeated for multiple times until model converges.
We conduct experiments on three publicly available medical NER datasets.
The experimental results validate that our method can boost the performance of medical NER by leveraging the labeled data on different platforms for model training in a collaborative way, and at the same time remove the need to directly exchange raw data among different platforms for better privacy protection.

The main contributions of this paper are summarized as follows:
\begin{itemize}
    \item We propose a \textit{FedNER} method based on federated learning to learn more accurate medical NER model from the labeled data of multiple medical platforms without the need to directly exchange the raw privacy-sensitive medical data among different platforms. 
    \item Different from original federated learning where all clients share the same model, in \textit{FedNER} we propose to decompose the medical NER model on each platform into shared and private modules to effectively leverage the knowledge from other platforms and at the same time capture the characteristics of the local data.
    \item We conduct extensive experiments on different benchmark datasets to verify the effectiveness of the proposed \textit{FedNER} method.
\end{itemize}

\section{Related Work}

Medical named entity recognition is a challenging research topic as it requires both understanding of texts and domain knowledge.
Both rule-based methods and statistical methods are proposed to tackle NER in the medical domain~\cite{dong2016multiclass,nadeau2007survey}.
For example, Embarek et al.~\shortcite{embarek2008learning} developed a rule-based tagging systems by capturing linguistic patterns, e.g., inflected form and lemma.
Other rule-based approaches involve some domain-specific knowledge bases or tools, such as MetaMap~\cite{aronson2001effective} and UMLS~\cite{odisho2019pd58}.
However, most of these rule-based methods require heavy effort and expertise to design effective rules.
Thus, statistical methods have also been widely adopted, ranging from SVM~\cite{isozaki2002efficient} to more recent neural methods~\cite{wang2018label,jain2015supervised}.
For instance, Xu et al.~\shortcite{xu2017bidirectional} proposed to use bi-directional LSTM to learn both character and word embeddings and CRF for label decoding.
Zhao et al.~\shortcite{zhao2019neural} proposed a joint learning approach for medical entity recognition and normalization.
It uses a character-level CNN to form word representations along with the pre-trained word embeddings and a Bi-LSTM to learn contextual representation of words.
One problem for these methods is the dependency on a large-scale and well-annotated corpus, facing a small corpora, their performances may degrade significantly.
However, the labeled data in one single medical platform is usually limited, and annotating a large-scale corpora is laborious and time-consuming.
Without sufficient labeled data, it is difficult for these deep learning based methods to achieve satisfactory performance.
Although different medical platforms may have some own labeled datasets, they cannot be directly aggregated since medical data is highly privacy-sensitive~\cite{sweeney2000simple}. 
Both uploading medical data from different platforms to a server and exchanging it between different platforms will cause high risk of privacy leakage.
Moreover, recent laws and regulations such as GDPR\footnote{https://gdpr-info.eu/} have enforced strict requirements on protecting the privacy of user data.

Recently, federated learning is proposed by McMahan et al.~\shortcite{mcmahan2017communication} to collectively train intelligent models from the locally stored data of massive users and remove the need to upload it to server to reduce the risk of privacy and security.
In federated learning, all user client share the same model which is coordinated by a central server.
Each client updates its local model with private data and transmits the local model update to the central server.
The server then aggregates received model updates from massive user clients, updates the global model and distributes the new model to each client for next-round training.
In federated learning, the raw data never leaves the user devices and only model updates are uploaded to server, which generally contain less information than the raw data.
Federated learning has been applied to a few NLP tasks to exploit the corpus from different sources in a privacy-preserving way~\cite{jiang2019federated,hardy2017private}.
For instance, Jiang et al.~\shortcite{jiang2019federated} proposed a federated topic modeling approach, which trains a unified high-quality topic model using data from multiple sensitive text corpus.
In existing federated learning methods, different clients usually share the same model, assuming that the private data of different clients share the same characteristics.
However, in medical NER, the entity types and annotation criteria of different medical platforms usually have significant difference.
Thus, in our \textit{FedNER} method we decompose the medical NER model on each medical platform into a shared module and a private module to leverage the shareable knowledge among different platforms and at the same time capturing the characteristics of the data on each platform.
\section{FedNER Method}



\begin{figure}[t]
	\centering
	\resizebox{0.4\textwidth}{!}{\includegraphics{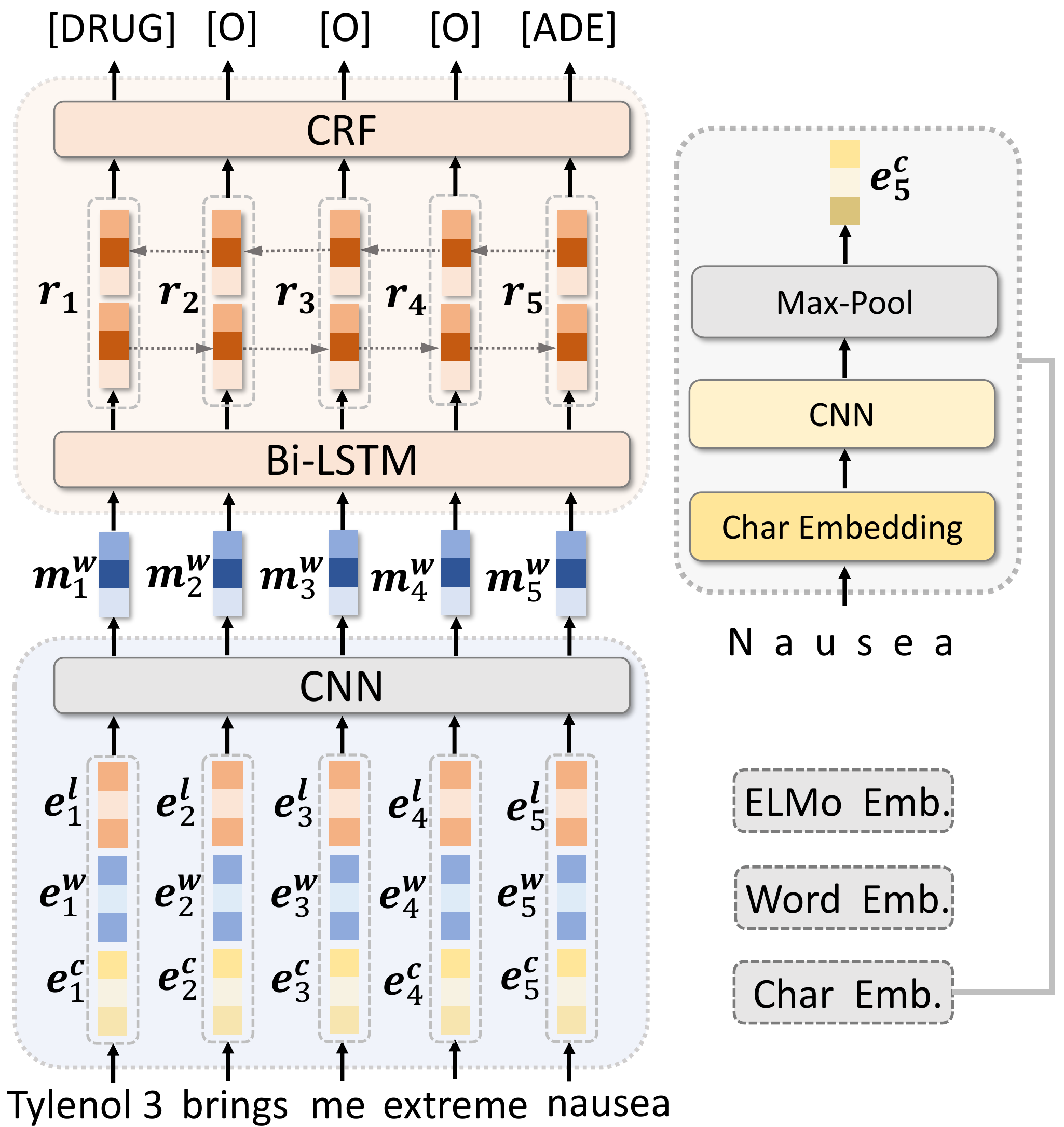}}
	\caption{Medical NER model.}
	\label{fig:model}
\end{figure}

\begin{figure*}[t]
	\centering
	\resizebox{0.7\textwidth}{!}{\includegraphics{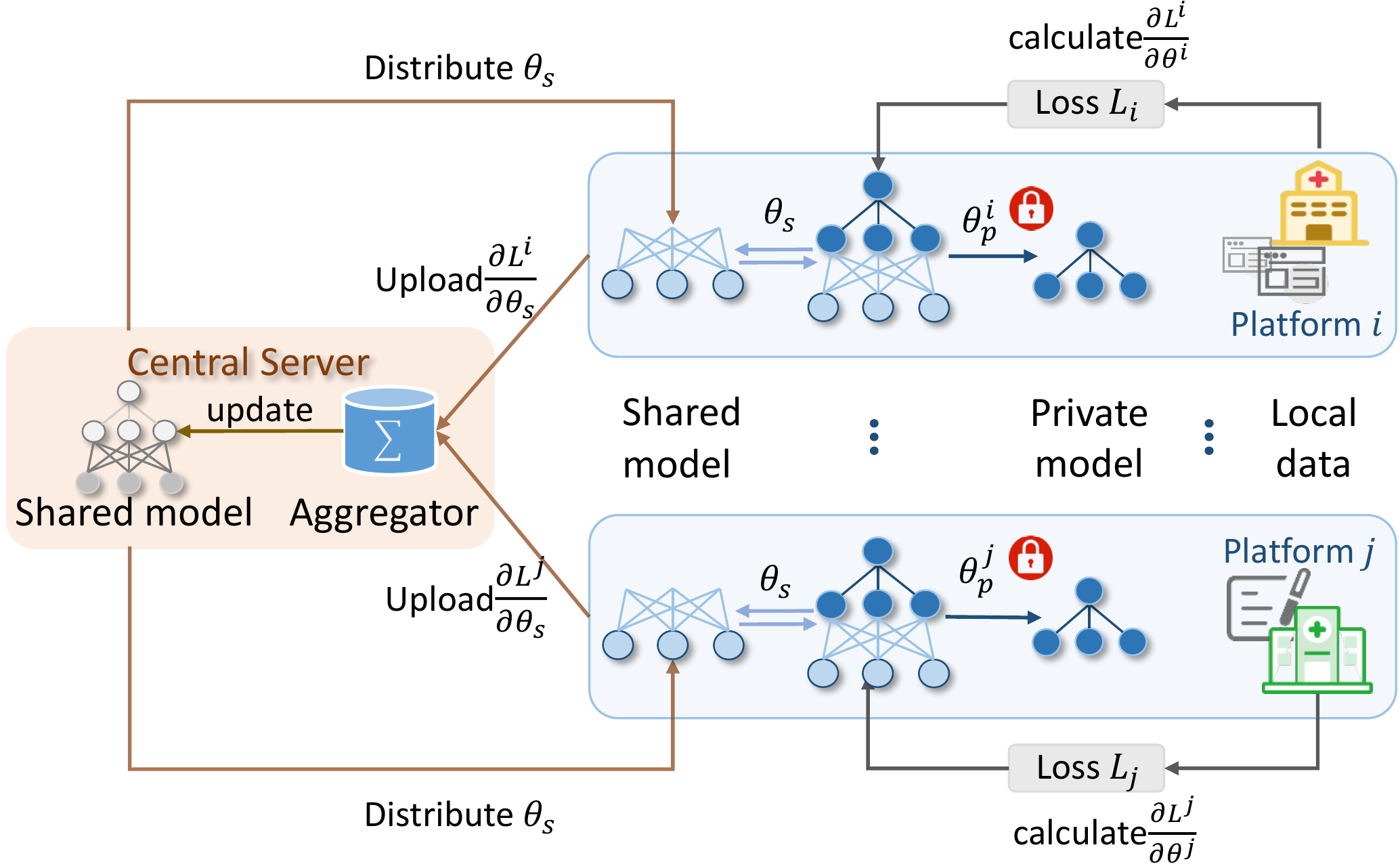}}
	\caption{The overall framework of FedNER.}
	\label{fig:framework}
\end{figure*}

In this section we first introduce the basic medical NER model used in our method.
Then we introduce the \textit{FedNER} framework for privacy-preserving medical NER model training with data from different medical platforms.

\subsection{Medical NER Model}

Following many existing works~\cite{xu2017bidirectional,zhao2019neural}, we formulate medical NER as a sequence labeling task. For example, given an input sentence ``Aspirin causes me a severe headache'', the medical NER model will output a tag sentence ``\footnotesize [DRUG] [O] [O] [O] [ADE] [ADE]\normalsize ''.\footnote{``ADE'' stands for adverse drug effect.}
In this section, we will introduce our medical NER model illustrated in Figure~\ref{fig:model}, which contains three sub-modules, i.e., \textit{word representation}, \textit{context modeling} and \textit{label decoding}.

The \textit{word representation} module incorporates three kinds of embedding to represent words, i.e, pre-trained word embedding, character-based embedding and language model embedding.
The pre-trained word embedding represents each word using a semantic vector.
Denote a sentence with $k$ words as $[w_1,w_2,...,w_k]$, it is converted to an embedded sequence $[\mathbf{e}_1^w,\mathbf{e}_2^w,...,\mathbf{e}_k^w]$ through the embedding matrix $\mathcal{M}_w \in \mathcal{R}^{D_w \times N_w}$, where $D_w$ is the dimension of the word embedding and $N_w$ is the vocabulary size.
However, they are insufficient due to the massive rare and out-of-vocabulary (OOV) words in the medical field.
Thus, we additionally model each word at a character level.
For a word $w_i$ with $m$ characters, we first use an embedding matrix $\mathcal{M}_c \in \mathcal{R}^{D_c \times N_c}$ to obtain the character-level embedding outputs, where $D_c$ is the dimension of the character embedding and $N_c$ is the number of characters.
The output character embedding of word $w_i$ is denoted as $[\mathbf{c}_1,\mathbf{c}_2,...,\mathbf{c}_m]$.
To model the relation between characters in a word, we apply a CNN layer to learn contextual representation of each character, the output sequence of the word $w_i$ is denoted as $[\mathbf{m}_1^c,\mathbf{m}_2^c,...,\mathbf{m}_m^c]$.
Then the contextual character sequence is sent to a max-pooling layer, and transformed to the final character-based embedding $\mathbf{e}_i^c$ for word $w_i$.
For a sentence of length $k$, the final character-based embedding of a sentence is denoted as $[\mathbf{e}_1^c,\mathbf{e}_2^c,...,\mathbf{e}_k^c]$.
Besides, since many words are context-dependent, we use the pre-trained ELMo\footnote{We also tried BERT but we found ELMo can achieve better performance.} language model to produce context-aware embeddings~\cite{Peters:2018}.
The language model embedding of a sentence is denoted as $[\mathbf{e}_1^l,\mathbf{e}_2^l,...,\mathbf{e}_k^l]$.
The final representation $\mathbf{e}_i$ of the word $w_i$ is the concatenation of the above three types of embedding, which is formed as:
$\mathbf{e}_i=[\mathbf{e}_i^w; \mathbf{e}_i^c; \mathbf{e}_i^l]$.

The \textit{context modeling} module utilizes two layers to enhance the word representations by capturing the dependency between words. 
The first layer is a word-level CNN, which aims to capture local context information~\cite{kim2014convolutional}.
Many medical entities are the short combination of several words, e.g., ``itchy scalp'' and ``restless leg syndrome''.
Thus, modeling relations between near neighbours may help better recognize entities.
Given an output sequence $[\mathbf{e}_1,\mathbf{e}_2,...,\mathbf{e}_k]$ from the \textit{word representation} module as input, the word-level CNN output is denoted as $[\mathbf{m}_1^w,\mathbf{m}_2^w,...,\mathbf{m}_k^w]$.
The second layer is a Bi-LSTM, which aims to model long-distance dependencies between words in both directions~\cite{huang2015bidirectional}.
Some descriptions of health condition have a relatively long span, and modeling only local contexts may be insufficient.
For instance, in the expression ``hair has been definitely falling out", the interaction between ``hair'' and ``falling out'' is essential for entity prediction.
Thus, we use a Bi-LSTM layer to model this kind of relationship.
The output of the Bi-LSTM layer is the contextual sequence $[\mathbf{r}_1,\mathbf{r}_2,...,\mathbf{r}_k]$.
By using a combination of CNN and LSTM, both local and global contexts can be taken into consideration.

The \textit{label decoding} module aims to decode word labels.
Neighbor labels usually have relatedness with each other in NER task.
Thus, we use the conditional random field (CRF) to jointly decode the optimal label chains by considering label dependencies.
The loss function of our NER model is formulated as:
\begin{equation}
    \mathop{\mathcal{L}}\limits_{\theta, \mathbf{y}, \mathbf{s}}=-\sum_{\mathbf{s} \in \mathcal{D}}\log p(\mathbf{y}|\mathbf{s}),
\end{equation}
where $\theta$ is all the trainable parameters in the NER model, $\mathcal{D}$ is the labeled training dataset, $\mathbf{s}$ is a word sequence and $\mathbf{y}$ is the corresponding label sequence.
\subsection{FedNER Framework}

\begin{algorithm}[t]
    \SetAlgoNoLine 
    \SetKwInOut{Input}{\textbf{Parameters}}\SetKwInOut{Output}{\textbf{dsds}} 
    \rm \textbf{Parameters:}\\
    \quad    The platform set \footnotesize $\mathcal{P}$\normalsize\;
    \quad The global model batch size \footnotesize $\mathcal{N}$\normalsize\;
    \quad   The training dataset \footnotesize $\mathcal{S}_i$ \normalsize of the $i_{th}$ platform\;
    \quad The learning rate $\alpha$.
    \BlankLine
    Initialize $\theta_s$ on the server\; 
    \BlankLine
    \Repeat
            {\text{converge}}
            {
                The server distributes $\theta_s$ to each platform\; 
                \For {each platform $i$ $\in$ \footnotesize $\mathcal{P}$ 
                \normalsize {\rm \textbf{in parallel}}}{
                  $\textit{\textbf{PlatformUpdate($\theta_s$, \footnotesize $\mathcal{N}$\normalsize, \footnotesize $\mathcal{S}_{i=1,2,...}$ \normalsize)}}$\;
                  Store the received $\frac{\partial \mathcal{L}^i}{\partial\theta_s}$\;
                }
                The server computes $\frac{\partial \mathcal{L}}{\partial\theta_s}$ using Eq.~(\ref{equ:agg})\;
                The server updates the model of the shared module $\theta_s \leftarrow \theta_s-\alpha \frac{\partial \mathcal{L}}{\partial\theta_s}$\;
            }
    \BlankLine
    \BlankLine
    \textit{\textbf{PlatformUpdate($\theta_s$, \footnotesize $\mathcal{N}$\normalsize, \footnotesize $\mathcal{S}_{i=1,2,...}$ \normalsize)}}:\\
    \begin{adjustwidth}{0.5cm}{0cm}
    
    Select a mini-batch of data \footnotesize $\mathcal{N}_i$ \normalsize from \footnotesize $\mathcal{S}_i$ \normalsize\;
    
    Compute $\frac{\partial \mathcal{L}^i}{\partial\theta_p^i}$ and $\frac{\partial \mathcal{L}^i}{\partial\theta_s}$ on \footnotesize $\mathcal{N}_i$ \normalsize\;
    
    $\theta_p^i \leftarrow \theta_p^i-\alpha \frac{\partial \mathcal{L}^i}{\partial\theta_p^i}$\;
    
    \rm \textbf{return} $\frac{\partial \mathcal{L}^i}{\partial\theta_s}$
    \end{adjustwidth}
    \caption{The framework of FedNER }
    \label{alg}
\end{algorithm}

In this section, we introduce our privacy-preserving approach for medical NER.
The framework of our approach is shown in Figure~\ref{fig:framework}.
In this framework, the server coordinates multiple clients for local model updating and global model sharing.
To be more specific, the clients here are different medical platforms, and train their local models with privately stored data. 
The central server monitors each platform for gradient aggregation and performs global model updating once it has collected gradients from all platforms.
Then it distributes the updated parameters of the shared model to each platform for next-round model training.
The overall learning framework of FedNER is illustrated in Algorithm~\ref{alg}.

In the training phase, each platform computes the model gradients using its locally stored data.
The data distribution across different platforms is non-I.I.D., and each platform keeps its data as private, only updating gradients obtained from the data training process.
Since the medical data stored in different platforms may have different characteristics and annotation criteria, sharing all model parameters between them may not be an optimal solution.
For example, some platforms may use the BIO tagging scheme while others may prefer the more complex BIOES tagging scheme.
Besides, some platforms mainly aim to find drug names and their corresponding dosages, while others may be more user-oriented, requiring the system to recognize user symptoms and adverse drug effects.
Thus, we propose to decompose the model into a shared module and a private module.
The private module consists of two top layers in our medical NER model, i.e,  Bi-LSTM and CRF, which aim to learn platform-specific context representations and label decoding strategies.\footnote{The partition strategy of the private module and shared module will be further discussed in Section~\ref{sec:paramshare}.}
We train the private module only with local data and exchange neither its parameters nor gradients.
The shared module consists of the other bottom layers in our NER model, such as the word-level CNN and all types of embedding.
Different from the private module, the shared one mainly aims to capture the semantic information in texts.
Since training of this shared module involves tuning embedding and filters for medical domain, a large-scale and well-annotated corpus is usually necessary.
However, the labeled data in one single platform may be insufficient and usually can not generalize to represent an overall non-I.I.D. data distribution.
Moreover, simply sharing the raw data among platforms may make up data shortage at the sacrifice of privacy.
Thus, we propose to share this model among all platforms in a federated learning framework.
This framework facilitates model training by leveraging the useful information of labeled data in different platforms.

Denote the set of platforms as $\mathcal{P}$ and the global model batch size as $\mathcal{N}$.
For the $i^{th}$ platform in $\mathcal{P}$, the training dataset is $\mathcal{S}_i$ and the loss function is denoted as $\mathcal{L}^i$.
In the beginning of each iteration, the $i^{th}$ platform will first select a mini-batch of training data $\mathcal{N}_i$ from $\mathcal{S}_i$, where $|\mathcal{N}_i|=\frac{|\mathcal{S}_i|}{\sum_{j\in\mathcal{P}}|\mathcal{S}_j|} \mathcal{N}$. 
Then the $i^{th}$ platform computes the gradients associated with parameters in the private and shared modules as $\frac{\partial \mathcal{L}^i}{\partial\theta_p^i}$ and $\frac{\partial \mathcal{L}^i}{\partial\theta_s}$.
The parameters of the private module are locally updated by $\theta_p^i=\theta_p^i-\alpha \frac{\partial \mathcal{L}^i}{\partial\theta_p^i}$,
where $\alpha$ is the learning rate.
Gradients of the shared module $\frac{\partial \mathcal{L}^i}{\partial\theta_s}$ are sent to a third-party central server for information sharing among different platforms.
Instead of directly sharing raw data, our approach only uploads gradients of the shared module, which generally contain less privacy-sensitive information.

The central server contains an aggregator and a globally shared model.
Here we assume the server belongs to one trusted third party, which means it will not make any vicious attack~\cite{bansal2019}.
In the beginning of each iteration, the server first monitors each platform for any possible gradient uploading.
Once it receives gradients from one platform, the server will store them for future aggregation.
When the server finishes receiving gradients from $|\mathcal{P}|$ platforms, the aggregator aggregates the locally-computed gradients from all platforms.
The aggregated gradients $\frac{\partial \mathcal{L}}{\partial\theta_s}$ are weighted summations of the received locally-computed gradients, which are formulated as:
\begin{equation}
\frac{\partial \mathcal{L}}{\partial\theta_s}=\sum_{i\in\mathcal{P}} \frac{|\mathcal{S}_i|}{\sum_{j\in\mathcal{P}}|\mathcal{S}_j|} \frac{\partial \mathcal{L}^i}{\partial \theta_s}.
\label{equ:agg}
\end{equation}
Since gradients from different platforms are aggregated together, the information of labeled data in each platform is harder to be inferred.
Thus, the privacy is well-protected.
The aggregator uses the aggregated gradients to update the parameters of a globally shared model stored on the central server by $\theta_s=\theta_s-\alpha \frac{\partial \mathcal{L}}{\partial\theta_s}$.
The updated globally shared model is then distributed to each platform to update their local shared module.
The process described above is repeated iteratively until the entire model converges.

In \textit{FedNER}, the medical NER model learning can benefit from incorporating the annotated information of the labeled data on different platforms, and the privacy is also well-protected by removing the need to exchange raw data directly among different platforms. 
\section{Experiments}

\subsection{Dataset and Experimental Settings}

\newcommand{\tabincell}[2]{\begin{tabular}{@{}#1@{}}#2\end{tabular}} 

\begin{table}[t]
    \centering
    \resizebox{0.49\textwidth}{!}{
    \begin{tabular}{c|c|c|c}
        \Xhline{1.2pt}
        \textbf{Dataset} & \textbf{\# Sentences} & \textbf{Entity types} & \textbf{\# Entities} \\
        \hline
        C{\scriptsize ADEC} & 7,597 & \tabincell{c}{Drug, ADE, Disease, \\Finding, Symptom}& 4,331\\
        \hline
        ADE Corpus & 4,484 & Drug, ADE, Dosage & 4,785\\
        \hline
        SMM4H & 2,213 & Drug, ADE & 1,209\\
        \Xhline{1.2pt}
    \end{tabular}}
    \caption{Statistics of the medical NER datasets.}
    \label{tab:dataset}
\end{table}

\begin{table*}[t]
	\centering
	\resizebox{\textwidth}{!}{
	\begin{tabular}{c|ccc|ccc|ccc|ccc|ccc|ccc}
    \Xhline{1.2pt}
	\multirow{3}{*}{\textbf{Model}} &\multicolumn{6}{c|}{\textbf{C{\scriptsize ADEC}}}&\multicolumn{6}{c|}{\textbf{ADE Corpus}}&\multicolumn{6}{c}{\textbf{SMM4H}}\\ 
			\cline{2-19}
			&\multicolumn{3}{c}{\textbf{Strict}}&\multicolumn{3}{c|}{\textbf{Relax}}& \multicolumn{3}{c}{\textbf{Strict}}&\multicolumn{3}{c|}{\textbf{Relax}}& \multicolumn{3}{c}{\textbf{Strict}}&\multicolumn{3}{c}{\textbf{Relax}}\\
            & \multicolumn{1}{c}{F1} & \multicolumn{1}{c}{Pre.} & \multicolumn{1}{c}{Rec.} & \multicolumn{1}{c}{F1} & \multicolumn{1}{c}{Pre.} & \multicolumn{1}{c|}{Rec.} & \multicolumn{1}{c}{F1} & \multicolumn{1}{c}{Pre.} & \multicolumn{1}{c}{Rec.} & \multicolumn{1}{c}{F1} & \multicolumn{1}{c}{Pre.} & \multicolumn{1}{c|}{Rec.} & \multicolumn{1}{c}{F1} & \multicolumn{1}{c}{Pre.} & \multicolumn{1}{c}{Rec.} & \multicolumn{1}{c}{F1} & \multicolumn{1}{c}{Pre.} & \multicolumn{1}{c}{Rec.}\\
			\hline
			CNN-CRF & 56.46 &55.29&57.68& 78.80 &80.09&77.55& 61.46 &59.84&63.17& 71.60 &65.21&79.38& 31.23 &37.94&26.54& 56.39&60.43&52.86 \\
			LSTM-CRF & 56.77 &56.13&57.42& 79.79 &77.77&81.92& 62.27 &61.13&63.45& 74.59 &67.92&82.71& 30.96 &36.58&\textbf{26.84}& 58.85&63.77&54.63 \\
			GRAM-CNN & 55.92 &55.08&56.79& 80.66 &81.16&80.17& 72.36 &68.78&76.33& 81.86 &75.00&90.10& 26.11 &28.99&23.75& 60.94&67.12&55.80 \\
			CNN-LSTM-CRF  & 59.30 &60.51&58.14& 81.46 &82.96&80.01& 73.04 &68.64&78.04& 83.50 &80.88&86.30& 27.65 &34.85&22.92& 61.99&79.89&50.64 \\
			S-LSTM-CRF  & 60.59 &61.34&59.86& 81.59 &79.71&83.56& 76.42 &77.09&75.76& 84.48 &81.59&87.58& 32.12 &40.77&26.50& 62.79&70.93&56.33 \\
			CNN-CLSTM-CRF & 60.14 &61.02&59.29& 81.94 &81.14&82.75& 78.83& 77.32& 80.40& 84.36 &83.81&84.92& 30.95 &38.84&25.72& 63.69&70.32&58.20\\
			\hline
			ELMoNER & 63.86& 63.03& \textbf{64.71}& 83.66& 84.05& 83.27& 81.62& 78.34& \textbf{85.19}& 87.44& 85.22& \textbf{89.79}& 31.51& 39.21& 26.34& 65.45& \textbf{80.32}& 55.23\\
			\hline
			FedNER & \textbf{65.16}& \textbf{66.45}& 63.92& \textbf{84.55}& \textbf{84.23}& \textbf{84.87}& \textbf{82.57}& \textbf{80.22}& 85.06& \textbf{88.90}& \textbf{88.39}& 89.42& \textbf{32.69}& \textbf{41.84}& 26.82& \textbf{67.81}& 74.27& \textbf{63.38} \\
    \Xhline{1.2pt}
	\end{tabular}}
	\caption{Results of different methods on medical NER.}
	\label{tab:baseline}
	\vspace{-0.05in}
\end{table*}

We experiment on three publicly available medical NER datasets, e.g., C{\scriptsize ADEC}~\cite{karimi2015cadec}, ADE Corpus~\cite{gurulingappa2012development} and SMM4H~\cite{weissenbacher2019overview}. 
The detailed information of these datasets is listed in Table~\ref{tab:dataset}.
Among all three datasets, there are 341 overlapping entities between them.
For each dataset, we randomly sample 80\% of sentences as training data, and the rest as testing data.
For word embedding, we use the pretrained Glove embedding~\cite{pennington2014glove}, which has a dimension of 300.
The dimension of the randomly initialized character embedding is 100.
The convolution layers of character-level and word-level CNN have 200 filters, with a kernel size of 3.
The Bi-LSTM layer has 2$\times$200 hidden states.
Adam is chosen as the optimizer with an initialized learning rate of 0.001.
We use dropout strategy to mitigate overfitting, the dropout rate is set to 0.2.
The aggregated number of gradients $\mathcal{N}$ is 64 in each interaction.
Following previous work~\cite{abacha2011medical}, we use the BIO tagging scheme.
We independently repeat each experiment 10 times and report the average strict F1 and relax F1 scores.
Under strict F1 evaluation, entity spans are considered correct only if position indices exactly match the gold annotations. 
Under relax F1 evaluation, only an overlap between the the range of predicted positions and gold annotations is needed. 

\subsection{Experimental Results}

We conduct experiments to compare the performance of our \textit{FedNER} method with several baseline NER methods, including: 
(1) CNN-CRF~\cite{collobert2011natural}, using CNN to learn word representations and CRF to decode labels;
(2) LSTM-CRF~\cite{habibi2017deep}, using a Bi-LSTM to learn word representations;
(3) GRAM-CNN~\cite{zhu2017gram}, using CNN to learn both character and word representations and CRF for label decoding.
(4) CNN-LSTM-CRF~\cite{ma2016end}, using CNN to learn character representations and LSTM to learn word representations;
(5) S-LSTM-CRF~\cite{lample2016neural}, a variant of LSTM-CRF that uses a stacked Bi-LSTM for word representation;
(6) CNN-CLSTM-CRF~\cite{shen2017deep}, a variant of CNN-LSTM-CRF, which uses a combination of CNN and LSTM to learn word representations;
(7) ELMoNER, the medical NER model introduced in Section 3.1, which is trained on single platform.
The results are summarized in Table~\ref{tab:baseline}.

We have two main findings from the results.
First, compared with other baseline NER methods, \textit{ELMoNER} can achieve better performances. 
This is because \textit{ELMoNER} learns word representation at both word level and character level.
By learning a character-level word representation, the model may better handle out-of-vocabulary medical terminologies by looking at their character contexts.
Besides, \textit{ELMoNER} captures both local and long-term context information by using a combination of CNN and Bi-LSTM networks.
Furthermore, it also utilizes context-aware word representations generated by the pre-trained language model ELMo to enhance representations of words.

Second, our \textit{FedNER} method can consistently outperform other methods on medical NER.
This is because the labeled data in a single medical platform is usually limited and insufficient to train an accurate NER model, and the datasets from different medical platforms cannot be exchanged due to the privacy sensitivity.
Different from the baseline methods which are trained on the data of a single medical platform, our \textit{FedNER} method can leverage the labeled data from different medical platforms in a privacy-preserving way to learn the shareable NER knowledge and alleviate the data sparsity problem.
Thus, our \textit{FedNER} method can achieve better performance on medical NER.

\subsection{Influence of Training Data Size}

\begin{figure}[t]
	\centering
	\resizebox{0.45\textwidth}{!}{\includegraphics{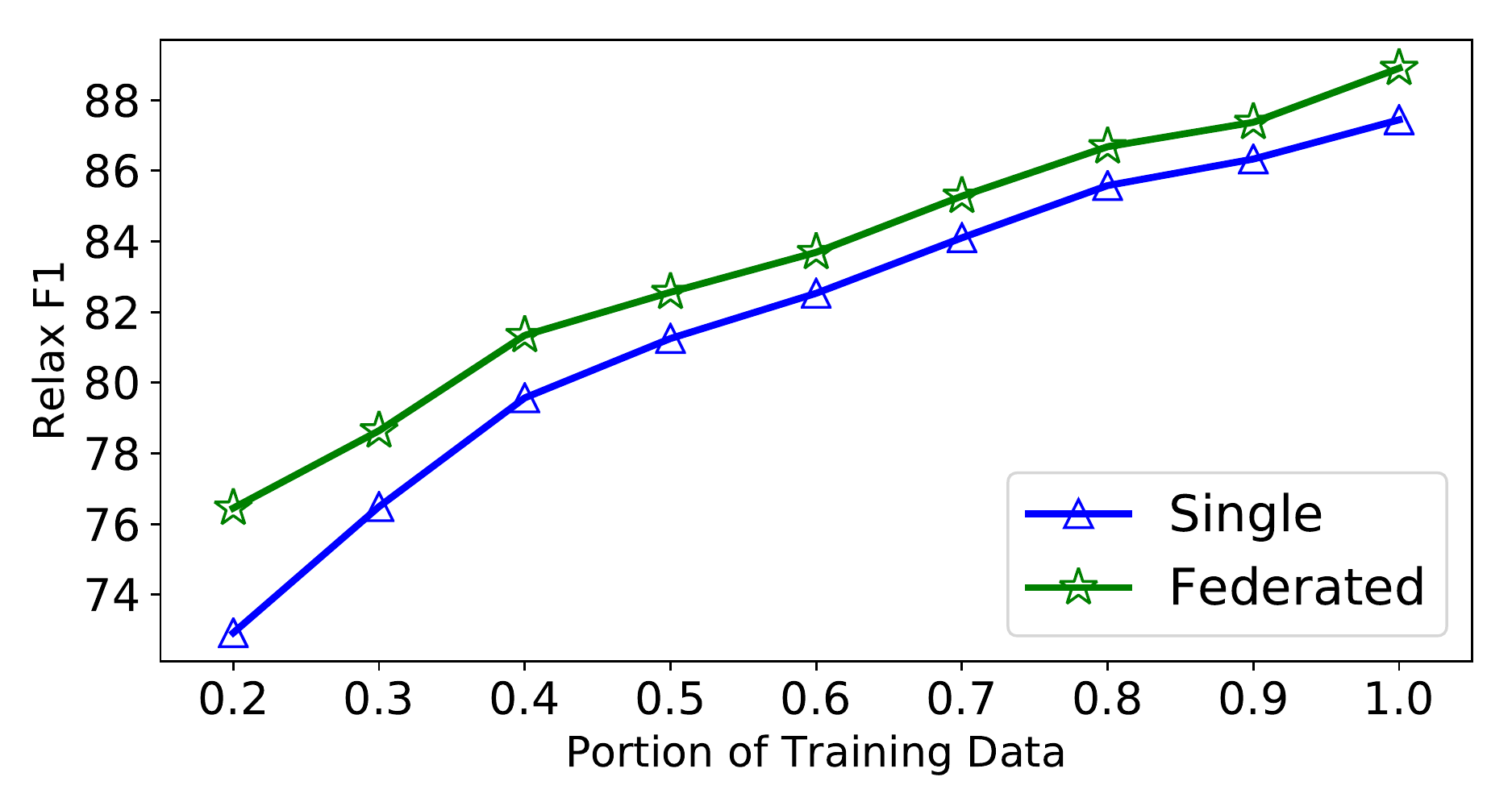}}
	\caption{Influence of training data size.}
	\label{fig:data}
\end{figure}

Next, we explore whether the proposed \textit{FedNER} can effectively handle the data scarcity problem on each platform by leveraging the useful data of different platforms.
We randomly select different ratios of data for model training, and due to space limit we only show the results on ADE Corpus dataset in Figure~\ref{fig:data}.
We find that compared with training model on the data of a single platform, the \textit{FedNER} can train more accurate medical NER model by leveraging the useful information from multiple platforms.
In addition, as the size of labeled data on each platform decreases, i.e., the data scarcity problem in single platform in more serious, and the performance improvement of \textit{FedNER} over single-platform training becomes more significant.
These results indicate that \textit{FedNER} can effectively leverage the useful information on different platforms to train more accurate medical NER model and alleviate the data scarcity problem on a single platform.

\subsection{Model Decomposition Strategy}

\begin{figure}[t]
	\centering
	\resizebox{0.45\textwidth}{!}{\includegraphics{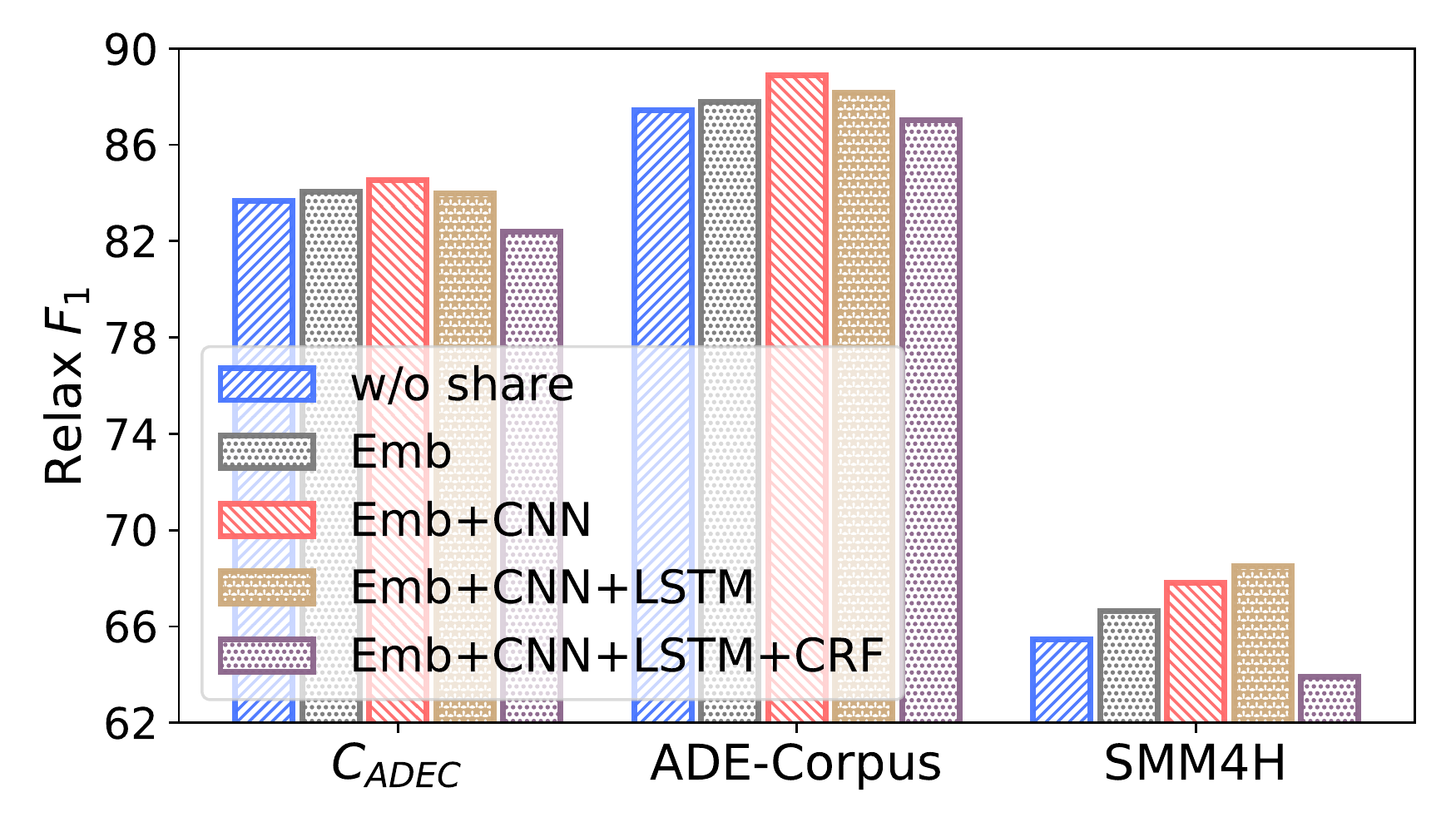}}
	\caption{\textit{FedNER} with different model decomposition strategies.}
	\label{fig:compare}
\end{figure}
\label{sec:paramshare}

In \textit{FedNER} we decompose the medical NER model into a shared module and a private module.
Next we explore the influence of different model decomposition strategies on the performance of \textit{FedNER}.
The results on ADE Corpus dataset are shown in Figure~\ref{fig:compare} and the results on other datasets show similar patterns.
We find that if the module is not shared, the performance is sub-optimal, since the shareable knowledge among different platforms is not exploited at all, and the labeled data on a single platform is insufficient to train an accurate enough model.
However, if all platforms share the same model, the performance is also not optimal.
This happens because the data on different platforms usually has different characteristics such as entity types and annotation criteria, which cannot be captured if we constrain different platforms to share exactly the same model.
These results validate the effectiveness of our strategy in decomposing the neural medical NER model into a shared module to learn the general and shareable knowledge for NER from multiple platforms, and a private module to capture the local data characteristics.

\subsection{Influence of Overlapped Entity Number}

\begin{figure}[t]
	\centering
	\resizebox{0.46\textwidth}{!}{\includegraphics{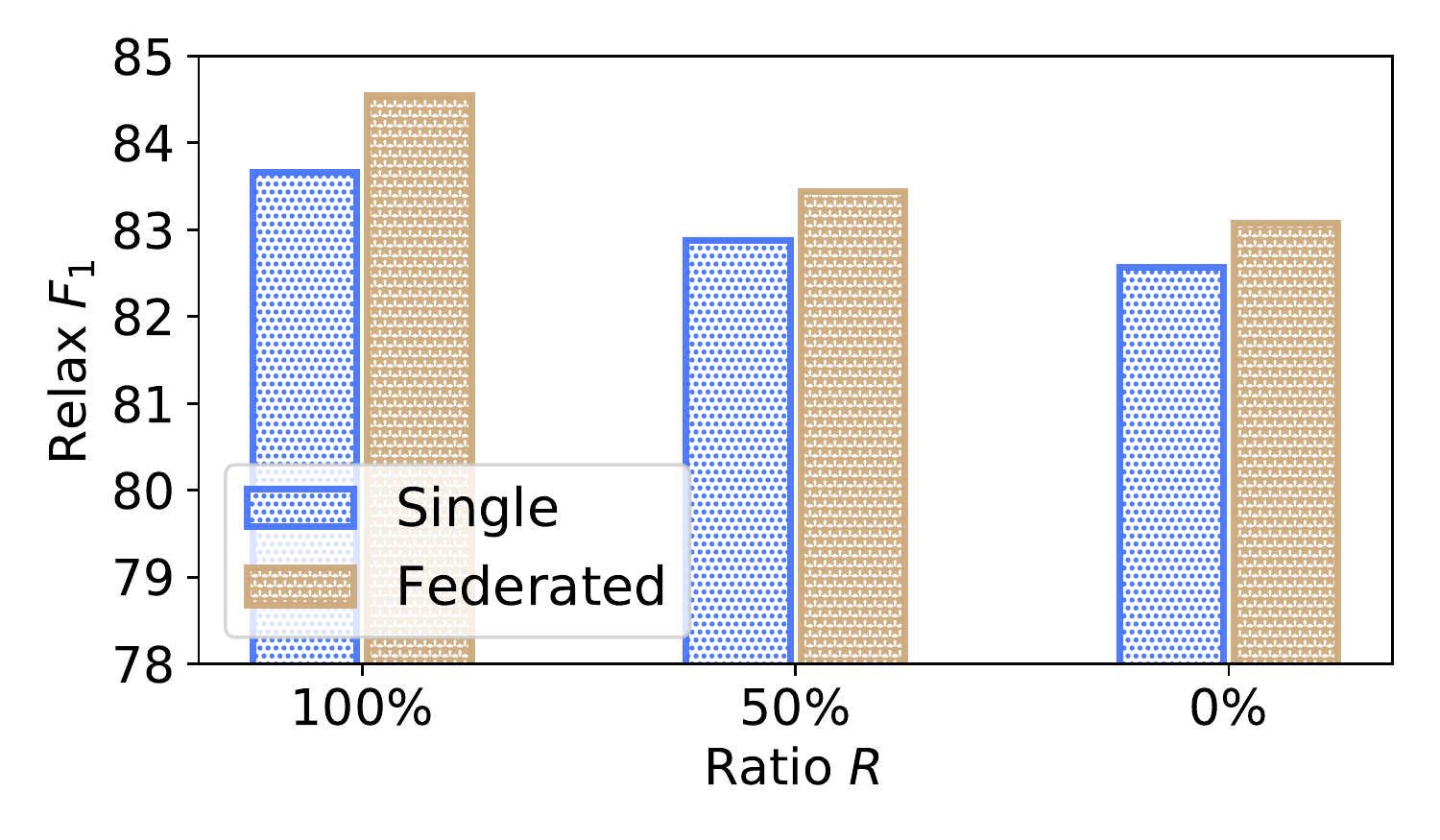}}
	\caption{Performance of \textit{FedNER} with different mask ratios of overlapped entities on the C{\scriptsize ADEC} dataset.}
	\label{fig:entity}
\end{figure}

A natural assumption is that the performance improvement of \textit{FedNER} is probably brought by the overlapped entities in different platforms.
In this section we explore this assumption by randomly masking different ratios of overlapped entities in training data.
The experimental results are shown in Figure~\ref{fig:entity}.
We have two findings from the results.
First, when there are more overlapped entities, the performance improvement of \textit{FedNER} over single-platform training is more significant.
This is intuitive, since the knowledge of overlapped entities can be easily learned in \textit{FedNER} by leveraging the information of different platforms for model training.
Second, even all the overlapped entities are masked, \textit{FedNER} can still bring consistent performance improvement.
This result shows that our \textit{FedNER} can learn some generalized knowledge of NER from the data of different platforms, rather than only the knowledge of overlapped entities.

\subsection{Generalization of FedNER Framework}

\begin{figure}[t]
	\centering
	\resizebox{0.48\textwidth}{!}{\includegraphics{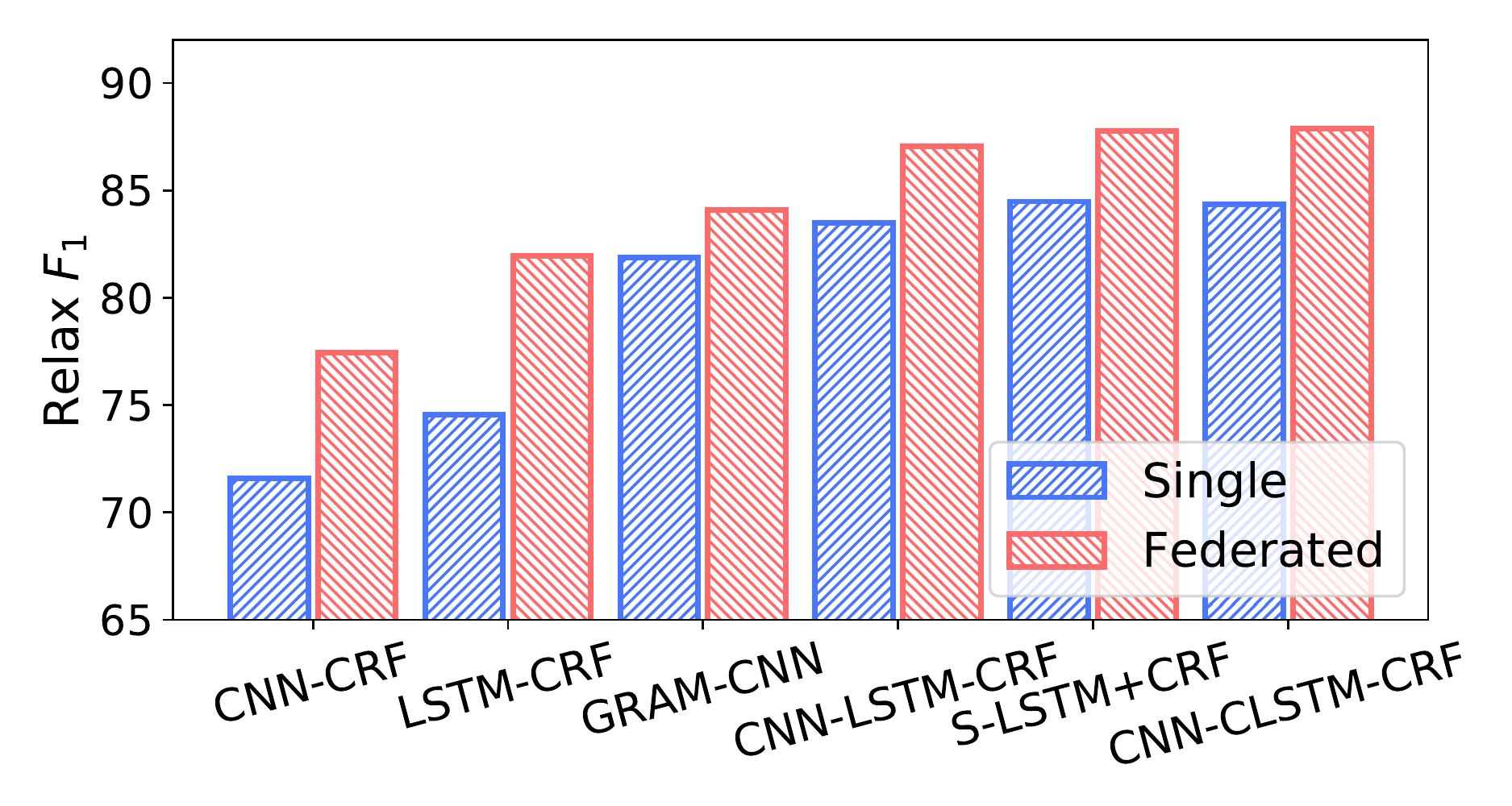}}
	\caption{Different medical NER methods under our FedNER framework on the ADE Corpus dataset.}
	\label{fig:general}
\end{figure}

To verify the generalization ability of the  \textit{FedNER} framework, we apply it to different existing medical NER methods to see whether they can benefit from leveraging the data on different platforms for model training  under our \textit{FedNER} framework.
We selected the best model decomposition strategy for each method.
The results on the ADE Corpus dataset are illustrated in Figure~\ref{fig:general}, and the results on other datasets show similar patterns.
We can see that all the existing medical NER methods compared here can achieve significant performance improvement under the \textit{FedNER} framework compared with single-platform training.
These results show that \textit{FedNER} is a general framework, and can help different medical NER methods to leverage the labeled data from different medical platforms in a privacy-preserving way to enhance the medical NER model training and alleviate data sparsity problem.

\section{Conclusion and Future Work}

In this paper, we propose a \textit{FedNER} method for medical NER.
It can train medical NER models by leveraging the labeled data on different platforms meanwhile removing the need 
to directly exchange the privacy-sensitive medical data among different platforms for better privacy protection.
We decompose the medical NER model in each platform into a shared module and a private module.
The private module is updated in each platform using the local data to model the platform-specific characteristics.
The shared module is used to capture the shareable knowledge among different platforms, and is updated in a server based on the aggregated gradients from multiple platforms.
It is further sent to each platform for next-round training.
Experiments on three benchmark datasets show our method can effectively improve the performance of medical NER by exploiting the useful information of multiple medical platforms in a privacy-preserving way.

In the future, we plan to strengthen the security guarantees of \textit{FedNER} by adopting the homomorphic encryption or local differential privacy techniques when gradients of the shared module are uploaded to the server.
In addition, we plan to apply \textit{FedNER} to other NER tasks with privacy-sensitive data on different platforms, such as financial text NER among different companies.

\bibliography{anthology,emnlp2020}
\bibliographystyle{acl_natbib}

\end{document}